\documentclass{article}

\PassOptionsToPackage{sort, compress, numbers}{natbib}

\usepackage[final]{neurips_2024_ml4ps}

\usepackage[utf8]{inputenc} 
\usepackage[T1]{fontenc}    
\usepackage{hyperref}       
\usepackage{url}            
\usepackage{booktabs}       
\usepackage{amsfonts}       
\usepackage{nicefrac}       
\usepackage{microtype}      
\usepackage{xcolor}         
\usepackage{amsmath}
\usepackage{relsize}
\usepackage{graphicx}

\title{Zephyr quantum-assisted hierarchical Calo4pQVAE for particle-calorimeter interactions}

\author{%
  Ian Lu \\
  TRIUMF, \\
  Vancouver, BC V6T 2A3, Canada\\
  \texttt{ilu@triumf.ca}
  \And
  Hao Jia \\
  Department of Physics and Astronomy, \\
  University of British Columbia, \\
  Vancouver, BC V6T 1Z1, Canada \\
  \texttt{haojia@phas.ubc.ca} \\
  \And
  Sebastian Gonzalez \\
  TRIUMF, \\
  Vancouver, BC V6T 2A3, Canada\\
  \texttt{bastigonzalez2000@gmail.com} \\
  \And
  Deniz Sogutlu \\
  TRIUMF, \\
  Vancouver, BC V6T 2A3, Canada\\
  \texttt{dsogutlu@triumf.ca} \\
  \And
  J. Quetzalcoatl Toledo-Marin\\
  TRIUMF, \\
  Vancouver, BC V6T 2A3, Canada\\
  Perimeter Institute
  for Theoretical Physics, \\
  Waterloo, ON,
  N2L 2Y5, Canada \\
  \texttt{jtoledo@triumf.ca} \\
  \And
  Sehmimul Hoque\\
  Faculty of Mathematics,\\
  University of Waterloo,
  ON, N2L 3G1, Canada \\
  Perimeter Institute for Theoretical Physics, \\
  Waterloo, ON, N2L 2Y5, Canada \\
  \texttt{s4hoque@uwaterloo.ca} \\
  \And
  Abhishek Abhishek\\
  Department of Electrical \\
  and Computer Engineering, \\
  University of British Columbia, \\
  Vancouver, BC V6T 1Z4, Canada \\
  \texttt{abhiabhi@student.ubc.ca}
  \And
  Colin Gay \\
  Department of Physics and Astronomy, \\
  University of British Columbia, \\
  Vancouver, BC V6T 1Z1, Canada \\
  \texttt{cgay@phas.ubc.ca} \\
  \And
  Roger Melko \\
  Perimeter Institute
  for Theoretical Physics, \\
  Waterloo, ON,
  N2L 2Y5, Canada \\
  Department of Physics
  and Astronomy,\\
  University of Waterloo, \\
  ON, N2L 3G1, Canada \\
  \texttt{rgmelko@uwaterloo.ca} \\
  \And
  Eric Paquet \\
  Digital Technologies Research \\
  Centre, National Research Council, \\
  1200 Montreal Road, \\
  Building M-50 Ottawa, \\
  ON, K1A 0R6, Canada\\
  \texttt{eric.paquet@nrc-cnrc.gc.ca} \\
  \And
  Geoffrey Fox \\
  University of Virginia, \\
  Computer Science and \\
  Biocomplexity Institute, \\
  994 Research Park Blvd, \\
  Charlottesville, \\
  VA, 22911, USA\\
  \texttt{gcfexchange@gmail.com}\\
  \And
  Maximilian Swiatlowski \\
  TRIUMF, \\
  Vancouver, BC V6T 2A3, Canada\\
  \texttt{mswiatlowski@triumf.ca}
  \And
  Wojciech Fedorko \\
  TRIUMF, \\
  Vancouver, BC V6T 2A3, Canada\\
  \texttt{wfedorko@triumf.ca}
}

\begin{document}

\maketitle

\begin{abstract}
  With the approach of the High Luminosity Large Hadron Collider (HL-LHC) era set to begin particle collisions by the end of this decade, it is evident that the computational demands of traditional collision simulation methods are becoming increasingly unsustainable. Existing approaches, which rely heavily on first-principles Monte Carlo simulations for modeling event showers in calorimeters, are projected to require millions of CPU-years annually—far exceeding current computational capacities. This bottleneck presents an exciting opportunity for advancements in computational physics by integrating deep generative models with quantum simulations. We propose a quantum-assisted hierarchical deep generative surrogate founded on a variational autoencoder (VAE) in combination with an energy conditioned restricted Boltzmann machine (RBM) embedded in the model's latent space as a prior. By mapping the topology of D-Wave's Zephyr quantum annealer (QA) into the nodes and couplings of a 4-partite RBM, we leverage quantum simulation to accelerate our shower generation times significantly.  To evaluate our framework, we use Dataset 2 of the CaloChallenge 2022. Through the integration of classical computation and quantum simulation, this hybrid framework paves way for utilizing large-scale quantum simulations as priors in deep generative models. 
\end{abstract}


\section{Introduction}

The High-Luminosity Large Hadron Collider (HL-LHC), expected to be operational by the end of this decade, will offer unprecedented opportunities to measure the Higgs boson properties, explore the Standard Model in greater depth, while also searching for physics beyond the Standard Model \cite{atlas2013physics} 
A critical component of this endeavor is the vast amount of data obtained from numerical simulations, which play a crucial role in both the design of future experiments and in the analysis of current ones. These simulations, done with Geant4 \cite{collaboration2003geant4, allison2016recent}, accurately describe the collisions at the Large Hadron Collider (LHC). But this comes at the price of being computationally intensive. These simulations describe the interactions between detectors and primary particles, but also account for the interaction with secondary particles produced as the primary particles interact with the detector material. Such is the case with calorimeters, which measure energy deposition from showers of secondary particles. 
Current projection for the HL-LHC run estimate millions of CPU-years per year \cite{atlas2022deep}. Simulating one single event with Geant4 in an LHC experiment requires approximately 1000 CPU seconds, with the calorimeter simulation being the most resource-intensive module \cite{rousseau2023experimental}. Through the generation of these showers, non-negligible computational resources are being employed in keeping track of these particles. Deep generative surrogates are being developed to model the particle-calorimeter interactions in the simulation pipeline, potentially reducing the overall time to simulate single events by several orders of magnitude. Examples of these are Generative Adversarial Networks \cite{de2017learning, paganini2018accelerating, paganini2018calogan}, which are now an integral part of the simulation pipeline \cite{atlas2020fast, aad2022atlfast3}. Similarly VAEs \cite{buhmann2021decoding, atlas2022deep, salamani2023metahep}, Normalizing Flows \cite{krause2021caloflow, buckley2024inductive}, Transformers \cite{favaro2024calodream}, Diffusion models \cite{ mikuni2024caloscore,kobylianskii2024calograph, liu2024calo} and combinations thereof \cite{amram2023denoising, madulacalolatent, hoque2023caloqvae, toledo2024conditioned}, where the last reference combines a VAE with a two-partite quantum annealer (QA). The framework combining VAE with QAs has also been used in different contexts \cite{winci2020path, dixit2021training}.


\begin{figure}[ht]
    \centering
    \includegraphics[width=0.99\linewidth]{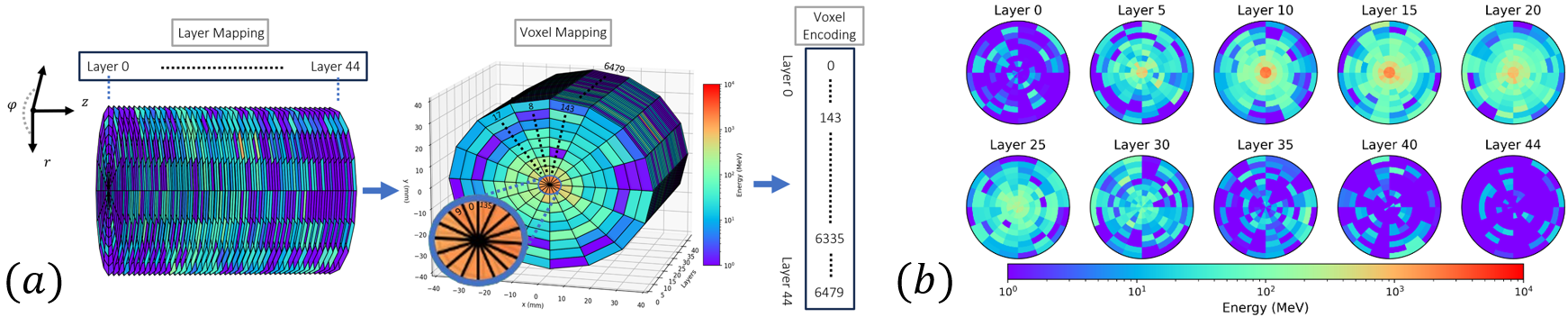}
    \caption{\textbf{(a)} Calochallenge dataset showers are voxelized using cylindrical coordinates $(r,\varphi,z)$. For any given event, each voxel value corresponds to the energy (MeV) in that vicinity. Each layer has 144 voxels composed of $16$ angular bins and $9$ radial bins. The data set is parsed onto a 1D vector with $6480$ voxels per each event. \textbf{(b)} Visualization of the voxels in an event in the dataset.}
    \label{fig:dataset}
\end{figure}

\section{Methods}

We illustrate our framework by using Dataset 2 of the CaloChallange-2022 \cite{calochallenge}. This dataset consists of 100k Geant4-simulated electron showers ranging from 1 GeV to 1 TeV incident particle energy, sampled from a log-uniform distribution. The voxelized detector is in the form of a concentric cylinder with 45 layers in the axial direction of which each layer is made up of an alternating collider-absorber, active (silicon) and passive (tungsten), material. Each layer consists of 144 voxels (volumetric pixels), 9 radially and 16 in the angular direction to yield a total of 45 x 16 x 9 = 6480 voxels in one event as shown in Fig. \ref{fig:dataset}. Each event has its corresponding incident particle energy as its label.
We preprocess the data similar to \cite{amram2023denoising}, except we omit the last step where the new variable is standardized. Instead we apply a shift to the logits to preserve the sparsity of the shower in the new variables, \textit{i.e.}, the new variable is zero whenever the voxel energy is zero.

\begin{figure}[ht]
    \centering
    \includegraphics[width=1.0\linewidth]{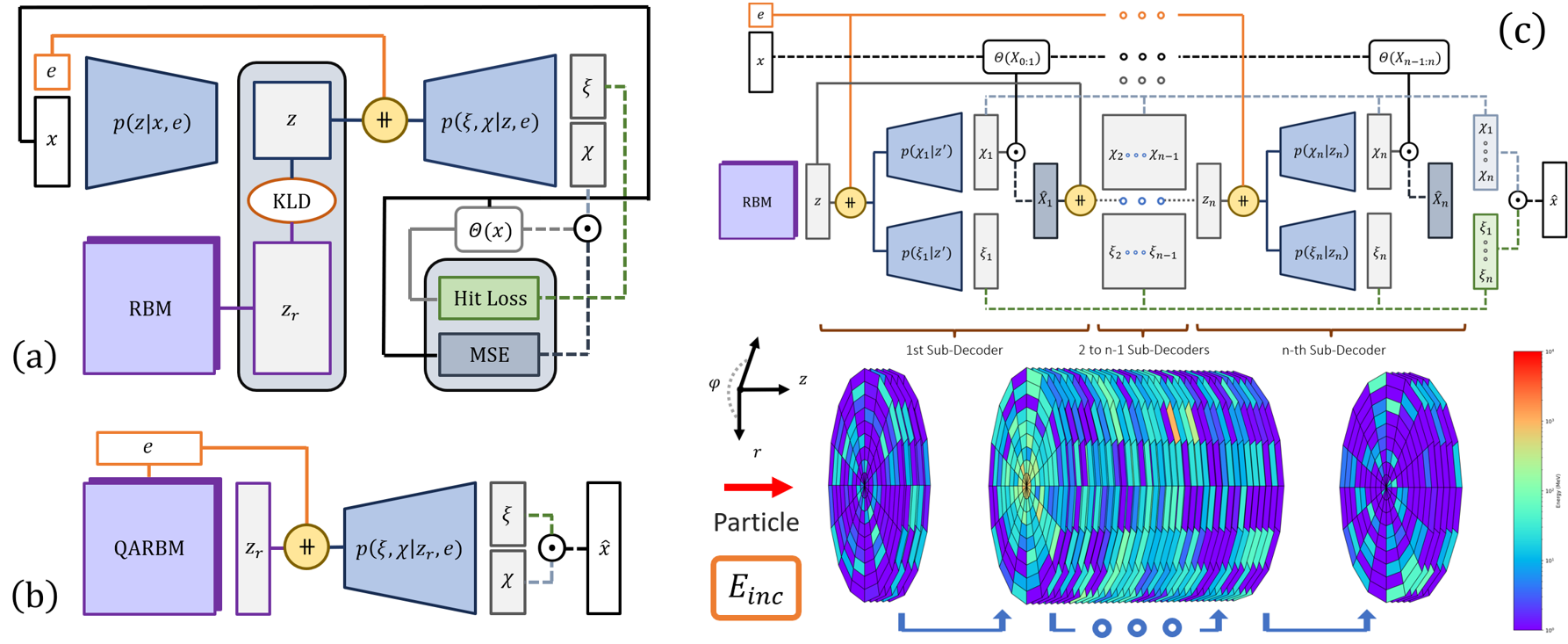}
    \caption{\textbf{(a)} Overview of Calo4pQVAE training architecture. Preprocessed voxels of a shower, $x$ and their corresponding incident energies, $e$ are inputted to the encoder. The encoder compresses the energy-conditioned shower into 4 partitions, of which 3 are generated from the hierarchies of the encoders and 1 is an encoded conditioning of the incident energy. The conditioned RBM is trained to learn these representations, then the  concatenation of the 4 partitions is the latent vector that gets passed through the hierarchical decoder, generating a hits and activations vector to reconstruct the shower.  \textbf{(b)} Once the model finishes training classically, the states of the trained RBM with an incidence energy conditioning is loaded onto D-Wave's Zephyr quantum annealer to sample a latent vector that is then passed through the hierarchical decoder to generate a shower. \textbf{(c)} The hierarchical decoder consists of 9 sub-decoders, each generating 5 layers to make up a total of 45 layers and conditions subsequent layers of the shower based on previous layers to simulate the physical propagation of particle scattering in the calorimeter through the evolution of the shower.}
    \label{fig:architectures}
\end{figure}

Our model is a variational autoencoder \cite{king2024computational} with a 4-partite conditioned restricted Boltzmann machine \cite{hinton2012practical} as the prior, as illustrated in Fig. \ref{fig:architectures} (a) . We used a hierarchical encoder composed of three sub-encoders. These hierarchy levels enforce couplings among latent units by introducing conditioning among latent nodes. In addition, these hierarchy levels introduce skip connections akin to residual networks \cite{he2016deep}. We feed the encoded sample from each of the three sub-encoder outputs to three of the partitions in the RBM, while the fourth partition is used to condition the RBM. The RBM condition parameter is the binarized incident particle energy of the event. The prior is the 4-partite restricted Boltzmann machine, where the connections between nodes mimic the Zephyr topology of D-wave's QA \cite{boothby2021zephyr}.  The encoded sample is then fed to the hierarchical decoder, as shown in Fig. \ref{fig:architectures} (c) where $n$ sub-decoders are allocated to generate $45 / n$ layers per sub-decoder. The hierarchical decoder conditions subsequent layers of the shower based on previous layers through hierarchies of auto regressive sub-decoders to simulate the physical propagation of particle scattering in the calorimeter during the evolution of a shower. The hierarchical decoder in Calo4pQVAE consists of 9 subdecoders, each generating 5 layers, making up the entire 45-layer voxelized shower. 
The decoder outputs a mask vector and an activation vector, and their Hadamard product yields the generated shower. We use the Gumbel trick \cite{maddison2016concrete} in our framework to generate both the encoded shower as well as to generate the output mask. Our code is publicly available and can be found here \cite{githubOurs}.

\begin{figure}[ht]
    \centering
    \includegraphics[width=0.99\linewidth]{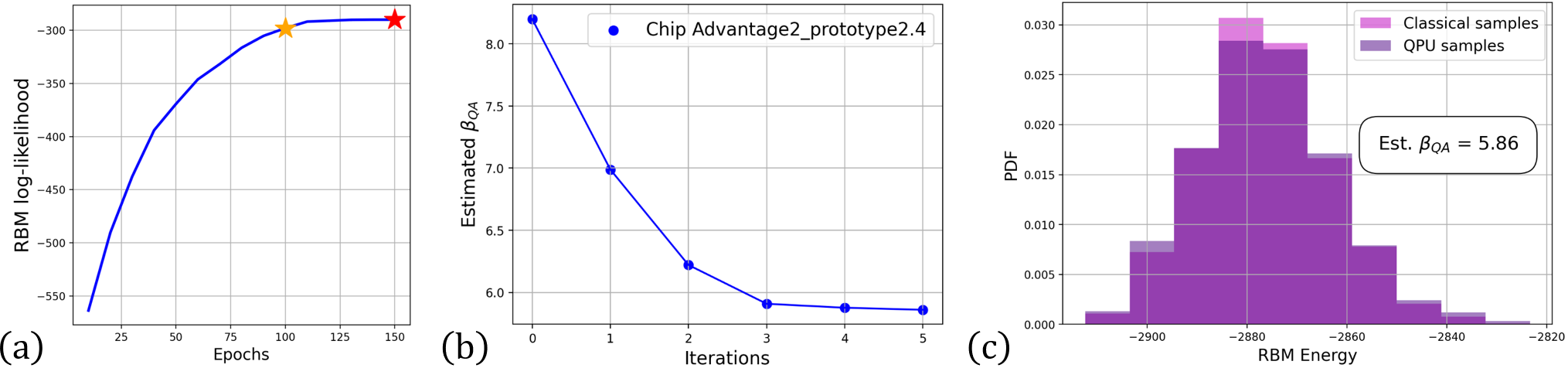}
    \caption{\textbf{(a)} Saturation of RBM log-likelihood vs epochs. Yellow star - freezing of encoder and decoder, Red star - completion of model training. \textbf{(b)} QA inverse temperature estimation \textit{vs} iterations. \textbf{(c)} RBM energy histogram for classical and QA samples.}
    \label{fig:qpus}
\end{figure}

\section{Results}
We trained our model classically for 100 epochs via the evidence lower bound (ELBO) function similar to \cite{hoque2023caloqvae}, set the number of Gibbs sampling steps for the RBM to 3000 and used contrastive divergence \cite{hinton2002training}. During the first 45 epochs we linearly annealed the activation function slope used in the Gumbel trick, from 5 to 500. Afterwards, we continued the training for another 45 epochs, afterwards we froze the encoder and decoder parameters and continued training the prior up to 150 epochs in total. The model was trained using NVIDIA RTX A6000. 
We validate our model using D-wave's Advantage2\_prototype2.3 for inference. 
It has been well documented how QAs can reach a freeze-out state \cite{amin2015searching}, akin to glass-forming melts under a fast quench \cite{toledo2016minimal}. Despite this, it has been shown that the distribution in this freeze-out state can be approximated with a Boltzmann distribution \cite{winci2020path}. We estimate the effective inverse temperature of the QA by means of a mapping with an attractive fix point at the QA's effective inverse temperature. This mapping is robust and converges faster than the method used in \cite{hoque2023caloqvae}. In Fig. \ref{fig:qpus} we show, \textit{(a)}, RBM log-likelihood vs epochs estimated via (reverse) annealed importance sampling \cite{salakhutdinov2008learning, burda2015accurate}, \textit{(b)}, a set of iterations to estimate the QA effective inverse temperature and, \textit{(c)}, the RBM energy histograms obtained from classical Monte Carlo and QA sampling.

\begin{figure}[ht]
    \centering
    \includegraphics[width=0.95\linewidth]{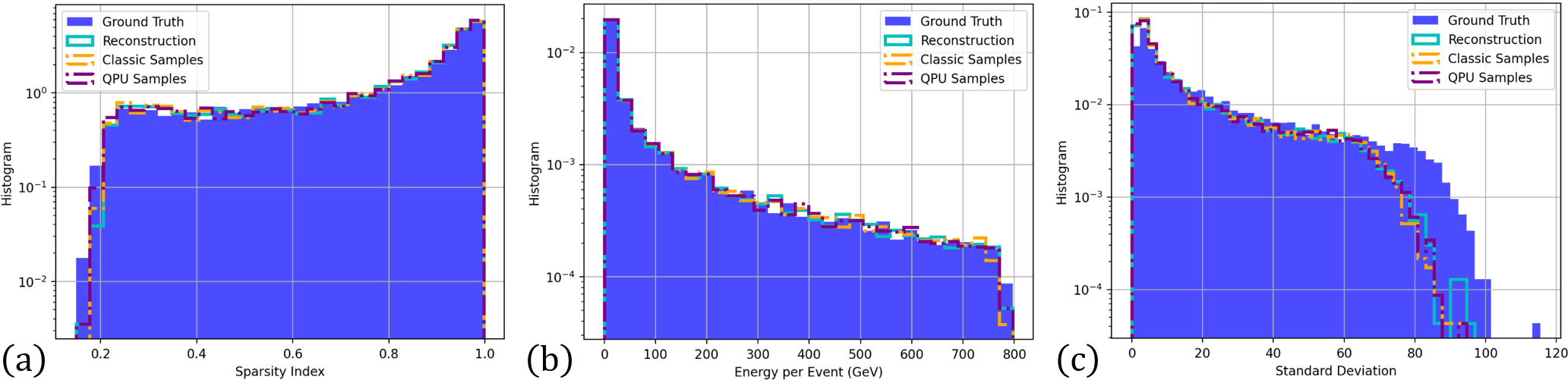}
    \caption{Normalized histograms comparing Geant4 simulated data (ground truth) and Calo4pQVAE's reconstruction, classically sampled synthetic data, and quantum annealed (Zeyphr) synthetic data for 10k events in: \textbf{(a)} sparsity index, ratio of non-hit voxels over all voxels in a shower, \textbf{(b)} energy per event, sum of all voxel energies in a shower, and \textbf{(c)} granularity, randomly shifted differences in voxel energies along angular and radial bins in a shower.}
    \label{fig:hists}
\end{figure}

\begin{figure}[ht]
    \centering
    \includegraphics[width=0.95\linewidth]{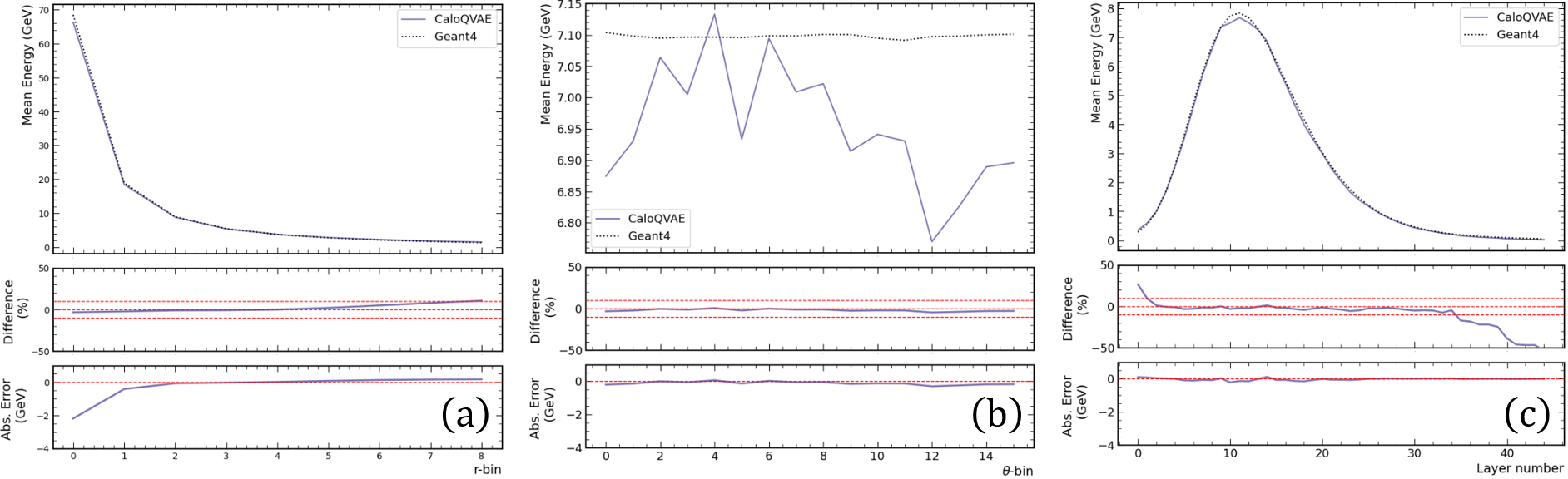}
    \caption{Solid vs dotted line plots comparing Geant4 simulated data and classical Calo4pQVAE's classically sampled synthetic data for 100k events. \textbf{(a)} Mean energy deposit per layer, \textbf{(b)} angular and \textbf{(c)} radial bin. Relative and absolute errors for each parameter are shown underneath each plot, respectively.}
    \label{fig:mean_energy}
\end{figure}

In Fig. \ref{fig:hists} we show the histograms for sparsity index (defined as the ratio between zero-energy voxels and total number of voxels), the energy per event and the shower standard deviation of the shower angular fluctuation, for ground truth, reconstruction, classical samples and QA samples. In Fig. \ref{fig:mean_energy} we compare the mean energy along the radial, angular and axial axis between the ground truth and our model. 
In Table \ref{tab:fpd_kpd_neurips} we present the Fr\'echet physics distance (FPD) and the Kernel physics distance (KPD) scores between our synthetic data and the Geant4 data, using the JetNet package \cite{jetlib}. These values are within the limits of the models analyzed in the CaloChallenge \cite{krause2024calochallenge}. There are additional metrics as part of the CaloChallenge that we will consider as an immediate continuation of this work.



\begin{table}[ht!]
    \centering
    \begin{tabular}{lcc}
        \toprule
        Model & FPD ($\times 10^{-3}$) & KPD ($\times 10^{-3}$) \\
        \midrule
        Calo4pQVAE & $1399.48 \pm 9.70$ & $15.94 \pm 1.02$ \\
        \bottomrule
    \end{tabular}
    \vspace{1mm}
    \caption{FPD and KPD values for Calo4pQVAE.}
    \label{tab:fpd_kpd_neurips}
\end{table}


\section{Conclusion}
In this paper we presented our 4-partite quantum-assisted deep generative model for calorimeter synthetic data generation. This framework provides competitive performance for simulating particle showers at the LHC experiments while running extremely quickly on D-wave quantum annealers. The quality of the synthetic data is average compared to other approaches \cite{amram2023denoising, mikuni2024caloscore}. This may be due in part to the sparse connectivity of the RBM, which mimics the QA connectivity. In addition to RBM connectivity, our framework could benefit from using attention layers similar to \cite{amram2023denoising}, which we leave for future work. Furthermore, Ref. \cite{toledo2024conditioned} presents an improved model that reaches KPD and FPD values of the order of $0.9 \cdot 10^{-3}$ and $450 \cdot 10^{-3}$, making our framework competitive compared to the frameworks analyzed in the CaloChallenge \cite{krause2024calochallenge}.

The generation time using GPU is dominated by the block Gibbs sampling steps to reach equilibrium. However, the number of steps to reach equilibrium is strongly dependent on training \cite{decelle2021equilibrium}. In our framework, we used $3000$ steps, which is more than typically used. Under these conditions, the generation time per event using GPU is roughly $500$ times faster than Geant4. Although the annealing time per sample using QPU is $20\;\mu s$, there is technical overhead. Under the previous conditions, the generation time per event using QPU is roughly one order of magnitude faster than using GPU. A more rigorous analysis is required in this comparison, due to the nuances involved in estimating the generation time using QPU and using GPU, and since our preliminary results indicate that they differ by one order of magnitude. We leave this for future work and reiterate that our framework is significantly faster than Geant4. In conclusion, our work on Calo4pQVAE demonstrates the utility of hybrid classical and quantum frameworks in generative AI. This hybrid framework opens new opportunities for leveraging large-scale quantum simulations as priors within deep generative models for high-energy physics and potentially beyond.

\begin{ack}
We gratefully acknowledge Jack Raymond, Mohammad Amin, Trevor Lanting and Mark Johnson for their feedback and discussions.
We gratefully acknowledge funding from the National Research Council (Canada) via Agreement AQC-002, Natural Sciences and Engineering Research Council (Canada), in particular, via Grants SAPPJ-2020-00032 and SAPPJ-2022-00020.
This research was supported in part by Perimeter Institute for Theoretical Physics. Research at Perimeter Institute is supported by the Government of Canada through the Department of Innovation, Science and Economic Development and by the Province of Ontario through the Ministry of Research, Innovation and Science. The University of Virginia acknowledges support from NSF 2212550 OAC Core: Smart Surrogates for High Performance Scientific Simulations and DE-SC0023452: FAIR Surrogate Benchmarks Supporting AI and Simulation Research. JQTM acknowledges a Mitacs Elevate Postdoctoral Fellowship (IT39533) with Perimeter Institute for Theoretical Physics.


\end{ack}

\bibliographystyle{unsrtnat}
\bibliography{references}

\end{document}